\documentclass[letterpaper, 10 pt, conference]{ieeeconf}
\IEEEoverridecommandlockouts    
\usepackage{subcaption}
 \usepackage{makecell} 

\usepackage{graphics}           
\usepackage{times}              
\usepackage{amsmath}            
\usepackage{amssymb}            
\usepackage{graphicx}
\usepackage{algorithm}
\usepackage[noend]{algpseudocode}
\usepackage{booktabs}
\usepackage{color}
\usepackage{listings}
\usepackage{subfiles}
\usepackage{hyperref}
\usepackage[nocompress]{cite} % sorts citations automatically, e.g., "[1],[2],[3]" instead of "[2],[3],[1]".
\definecolor{instructioncolor}{rgb}{.5,.5,.5}

\usepackage[font=small]{caption}

\def\figref#1{Fig.~\ref{#1}}
\def\tabref#1{Tab.~\ref{#1}}
\def\eqref#1{Eq.~(\ref{#1})}

\makeatletter
\usepackage{xspace}
\DeclareRobustCommand\onedot{\futurelet\@let@token\@onedot}
\def\@onedot{\ifx\@let@token.\else.\null\fi\xspace}

\def\etal{{et al}\onedot}
\makeatother

\def\etalcite#1{\etal~\cite{#1}}

\usepackage{array}
\newcolumntype{L}[1]{>{\raggedright\let\newline\\\arraybackslash\hspace{0pt}}m{#1}}
\newcolumntype{C}[1]{>{\centering\let\newline\\\arraybackslash\hspace{0pt}}m{#1}}
\newcolumntype{R}[1]{>{\raggedleft\let\newline\\\arraybackslash\hspace{0pt}}m{#1}}

\usepackage{comment}
\title{\LARGE \bf AgriGen: Large-Scale Scene Generation Framework for Photorealistic Agricultural Robotics Simulation}

\author{Utkarsh Bajpai\textsuperscript{1,*} \and  Serge Tleiji\textsuperscript{2,*} \and C\'{e}dric Pradalier\textsuperscript{1} \and St\'{e}phanie Aravecchia\textsuperscript{1}% <-this % stops a space
  \thanks{\textsuperscript{1} Georgia Tech-CNRS IRL2958, France }
  \thanks{\textsuperscript{2} Notre Dame University, Lebanon; work done during internship at \textsuperscript{1}}
  \thanks{\textsuperscript{*} Equal Contribution }
  \thanks{This work was supported by the European Union’s Horizon Europe research and innovation programme under the Marie Skłodowska-Curie Actions AIGreenBots Doctoral Network, under grant agreement ID 101169330.
  } 
  \thanks{Correspondance: \texttt{utkarsh.bajpai@georgiatech-metz.fr}}%
}

\begin{document}
\thispagestyle{empty}
\pagestyle{empty}
\maketitle

%%%%%%%%%%%%%%%%%%%%%%%%%%%%%%%%%%%%%%%%%%%%%%%%%%%%%%%%%%%%%%%%%%%%%%%%%%%%%%%%
\begin{abstract}
  Agricultural robotics is advancing rapidly, yet progress remains constrained by limited field access, lack of control over field conditions, geographic variability, and seasonal crop cycles. These factors make it difficult and costly to acquire diverse agricultural datasets, resulting in limited evaluation and reduced system robustness. While other robotics domains have scaled learning and evaluation through high-fidelity simulation, agricultural robotics still lacks comparably capable tools. In this paper, we present a ROS-integrated framework, built on Isaac Sim, for large-scale procedural generation of agricultural environments. The framework supports photorealistic rendering, physics simulation, and domain randomization at scales relevant to robotics research, with built-in support for row crops, orchards, and vineyards and straightforward extensibility to additional crop categories. 
  \\[1em]
  \hypersetup{hidelinks}  
  \noindent Project Page: \url{https://baj31415.github.io/agrigen/}
\end{abstract}

%%%%%%%%%%%%%%%%%%%%%%%%%%%%%%%%%%%%%%%%%%%%%%%%%%%%%%%%%%%%%%%%%%%%%%%%%%%%%%%%
\section{Introduction}
\label{sec:intro}
Agricultural robots require robust evaluation to validate performance under a diverse range of crop, terrain geometry, terrain type, illumination conditions and find corner cases. This typically requires physical access to robots and agricultural fields and control over field parameters such as layout, crop density, and weed distribution. In addition, agricultural robots that rely on data-driven methods require large amounts of labeled training data, including semantic and instance annotations. This need is particularly acute as deep learning is increasingly used for agricultural robotics tasks such as phenotyping, perception, navigation, and planning~\cite{Sivakumar2021rss}. In the broader robotics context, the challenges of data scarcity have increasingly been addressed through high-fidelity photorealistic simulators, which have played a central role in recent advances in robotics. Several domain specific simulators~\cite{savva2019habitat,kolve2017ai2thor,xiazamirhe2018gibson,dosovitskiy2017carla,shah2018airsim} have been proposed. Habitat~\cite{savva2019habitat} and Gibson~\cite{xiazamirhe2018gibson} target household robotics with diverse photorealistic indoor environments, supporting tasks such as navigation, manipulation, and SLAM. CARLA~\cite{dosovitskiy2017carla} focuses on autonomous driving environments while AirSim~\cite{shah2018airsim} as well as recent Isaac Sim--based platforms~\cite{xu2024omnidrones,kulkarni2025ral}, provide simulation support for aerial robots.

Agricultural environments remain largely absent from large-scale simulation efforts as seen in other domains such as household robotics or autonomous driving. Nevertheless, simulation platforms have increasingly been explored in agriculture as virtual testbeds for algorithm development and evaluation. Early efforts, including SEARFS~\cite{emmi2013searfs} and subsequent Gazebo-based systems~\cite{tsolakis2019agros, liu2025icra}, established simulation as a useful tool for agricultural field-operation evaluation, but remained constrained by simplified environments, limited photorealism, coarse assets, and small scene scale. Unity- and Unreal-based approaches have improved visual realism, such as in Agri-Ro5~\cite{cejudo2024agriro5} and the tabletop plant reconstruction simulator of Li~\etalcite{li2024agrisimunreal}, but remain subject to similar limitations in scale and diversity of environments~\cite{espejel2025comparison,mansur2023importance}. 

Most existing agricultural simulators rely on small, hand-crafted, simplified scenes that do not reflect the complexity of real fields. Even when photorealistic rendering and physics simulation are available, environmental diversity is typically bounded by fixed predefined 3D scenes reconstructed through photogrammetry or manually created by skilled artists. Existing platforms also typically support only a single crop type, partly due to the scarcity of open-access, high-fidelity vegetation assets. To the best of our knowledge, no existing framework supports agricultural-environment simulation at realistic field scales with diverse vegetation, including crops, weeds, grass, and varying terrain properties, for robotics research. To address this gap, we introduce AgriGen, a framework for large-scale agricultural robotics simulation. AgriGen enables simulation of robotic systems in settings that better capture the diversity, complexity, and variability of real agricultural field environments.
%
%%%%%%%%%%%%%%%%%%%%%%%%%%%%%%%%%%%%%%%%%%%%%%%%%%%%%%%%%%%%%%%%%%%%%%%%%%%%%%%%
\begin{figure*}[t]
  \centering
    \includegraphics[width=0.85\linewidth]{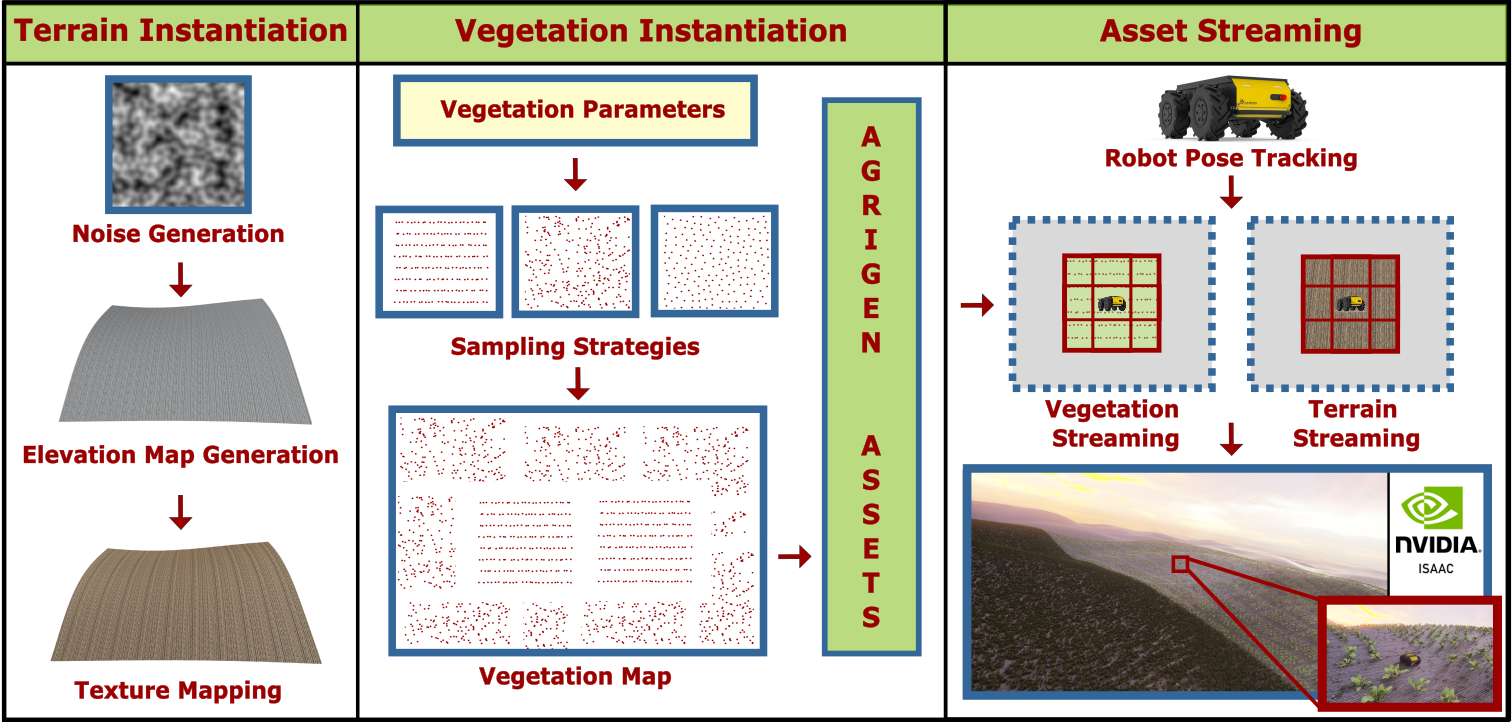}
  \caption{AgriGen generates large scale environments through procedural generation.}
      \vspace{-0.35cm}
  \label{fig:pipeline}
\end{figure*}
\begin{figure*}[ht]
  \centering
  \begin{subfigure}[t]{0.32\linewidth}
    \centering
    \includegraphics[width=0.9\linewidth]{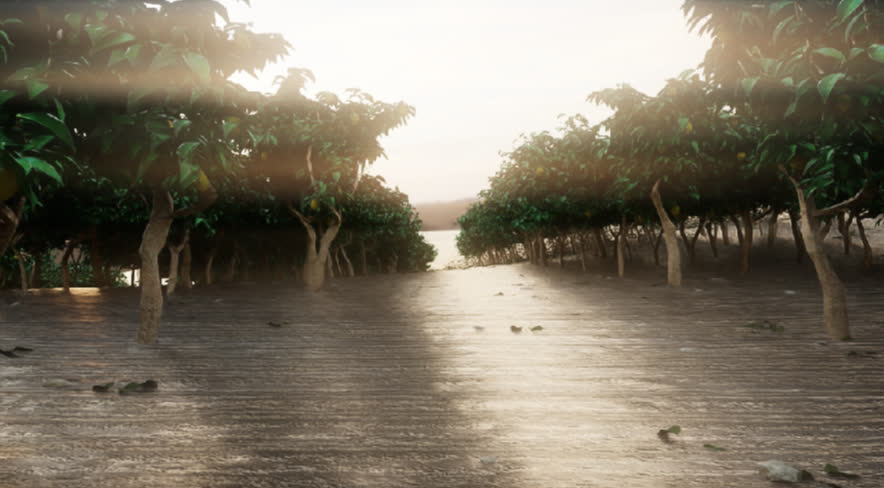}
    \caption{RGB}
  \end{subfigure}
  \hfill
  \begin{subfigure}[t]{0.32\linewidth}
    \centering
    \includegraphics[width=0.9\linewidth]{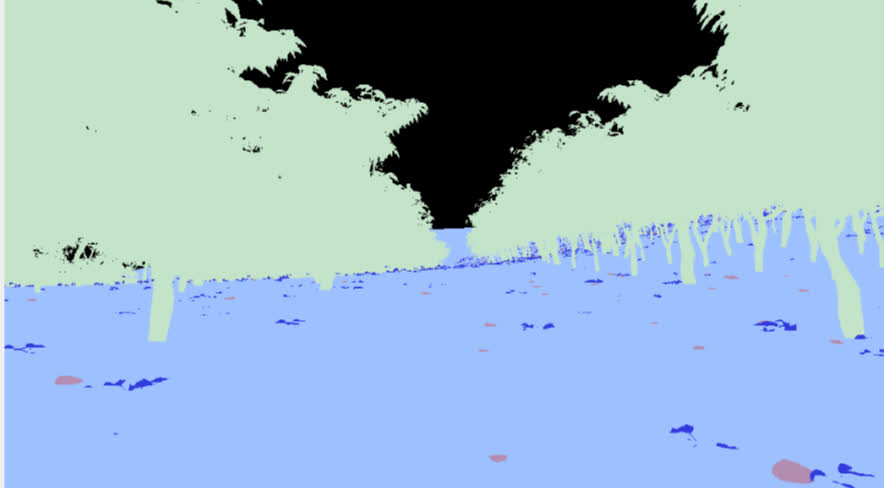}
    \caption{Semantics}
  \end{subfigure}
  \hfill
  \begin{subfigure}[t]{0.32\linewidth}
    \centering
    \includegraphics[width=0.9\linewidth]{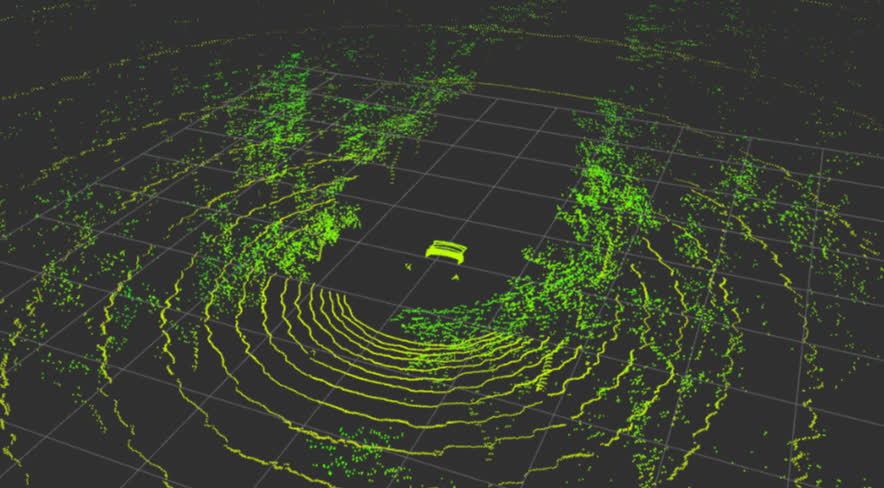}
    \caption{Pointcloud}
  \end{subfigure}
  \caption{Our framework enables photorealistic simulation of agricultural environments with ROS integration. The image illustrates simulated sensor data from the robot’s perspective, as visualized in RViz.}
  \label{fig:rviz_data}
    \vspace{-0.45cm}
\end{figure*}

\section{Our Approach}
\label{sec:approach}

AgriGen is a flexible procedural framework for generating large-scale agricultural environments in simulation. The generation process operates in three phases: the first two are the terrain and vegetation instantiation phases, where the simulation environment is structured. The third is the streaming phase, where terrain and vegetation assets are progressively loaded into the simulator at runtime. \figref{fig:pipeline} provides an overview of our method.

\subsection{Terrain Instantiation}
\label{sec:terrain_gen}

Agricultural environments often exhibit gently undulating terrain, ranging from flat plains to irregular hills and valleys. To model this variability, we generate an elevation field over the planar simulation domain using Perlin noise~\cite{perlin2002noise}. The field combines two spatial scales: a macro scale for broad hills and valleys, and a meso scale for smaller mounds and ridges. The resulting elevation map is converted into a triangular mesh representing the ground surface of the simulated environment. We then apply texture and bump mapping to assign color, material appearance, and fine-scale surface roughness such as soil irregularities. This produces diverse terrain with visually plausible structure and surface variation representative of real agricultural fields.
\subsection{Vegetation Instantiation}
\label{sec:asset_instantiation}
Crops form the primary structural elements of agricultural landscapes, but realistic environments also include grasses, weeds, shrubs, and rocks. To capture this diversity, we introduce a vegetation instancing framework that procedurally distributes assets using different sampling strategies and stores them on a vegetation map. The vegetation map is defined on the same planar domain as the elevation map.
\subsection{Asset Streaming}
\label{sec:asset_streaming}
Simulating a realistic crop field environment with thousands of individual vegetation assets is computationally intractable on consumer grade hardware. To this end, we adopt a streaming-based mechanism that enables efficient simulation of large-scale environments. We partition the planar domain into spatial tiles and load only those tiles into memory which are close to the robot’s current position and progressively unload tiles that are further away. Our streaming pipeline enables RTX rendering, dynamics simulation with PhysX physics, collision detection, as well as publishing of sensor data such as camera and LIDAR with ROS 2 integration on every simulation timestep.
\section{AgriGen as an agricultural robotics simulator}
\label{sec:exp}
We present our experiments to demonstrate the capabilities of our procedural scene generation framework for photorealistic agricultural robotics simulation. The results of our experiments also support our key claims, which are:
(i) Our ROS-enabled framework, built on Isaac Sim, enables the simulation of large-scale procedurally generated environments for agricultural robotics tasks.
(ii) Our pipeline natively supports the simulation of row crops, orchards and vineyards and is extensible to other crop categories. We release an asset dataset composing of 65 3D models of vegetation including crops, weeds, grass and shrubs.
(iii) Our framework enables the domain randomization of photorealism and physics properties in the same pipeline. We support randomization of terrain geometry, terrain type, asset type, asset geometry, asset distribution and lighting.

\begin{figure}[htb]
  \centering
  % First row
  \begin{subfigure}[t]{0.45\linewidth}
    \centering
    \includegraphics[width=\linewidth]{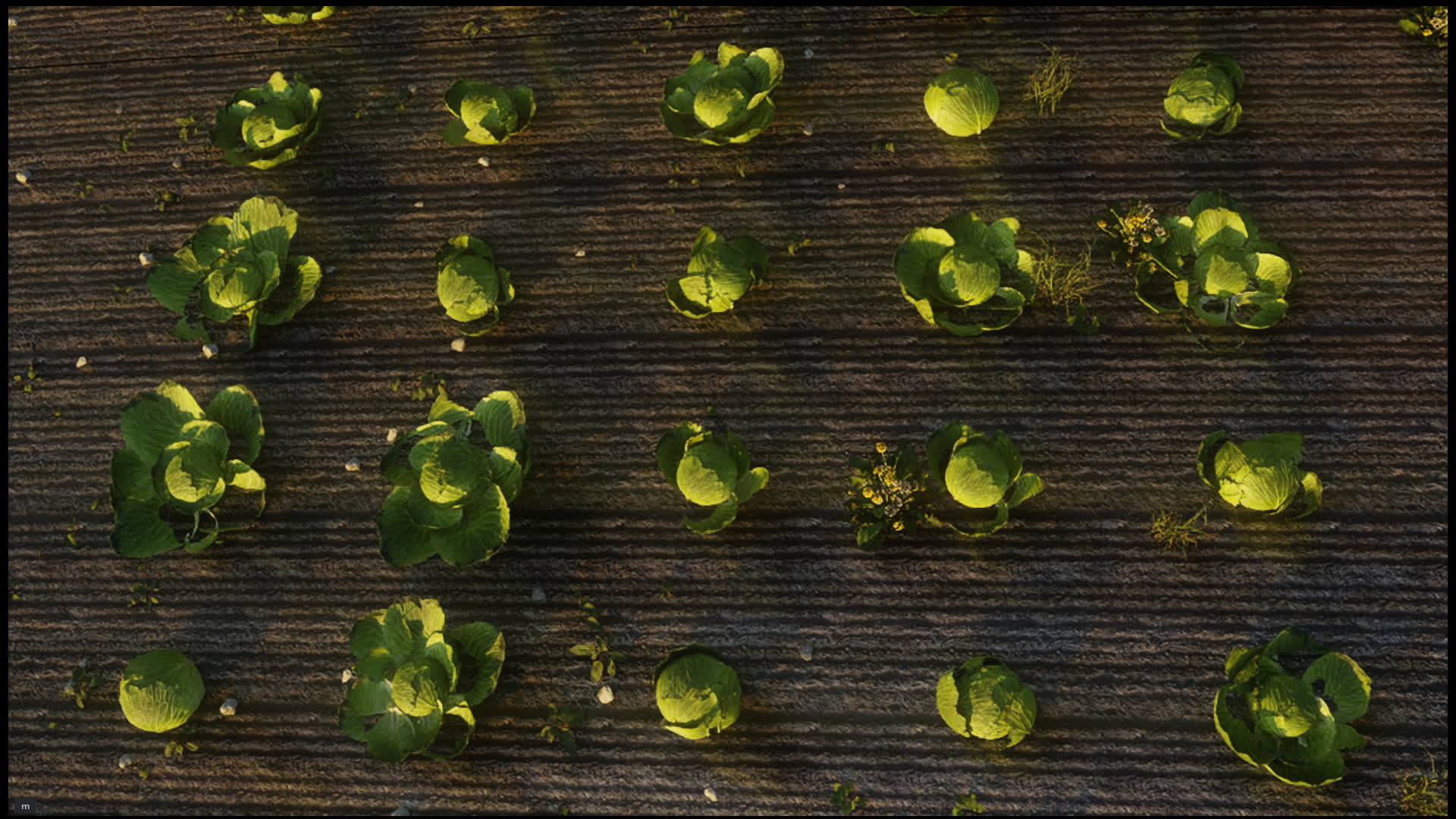}
    \caption{RGB}
  \end{subfigure}
  \hfill
  \begin{subfigure}[t]{0.45\linewidth}
    \centering
    \includegraphics[width=\linewidth]{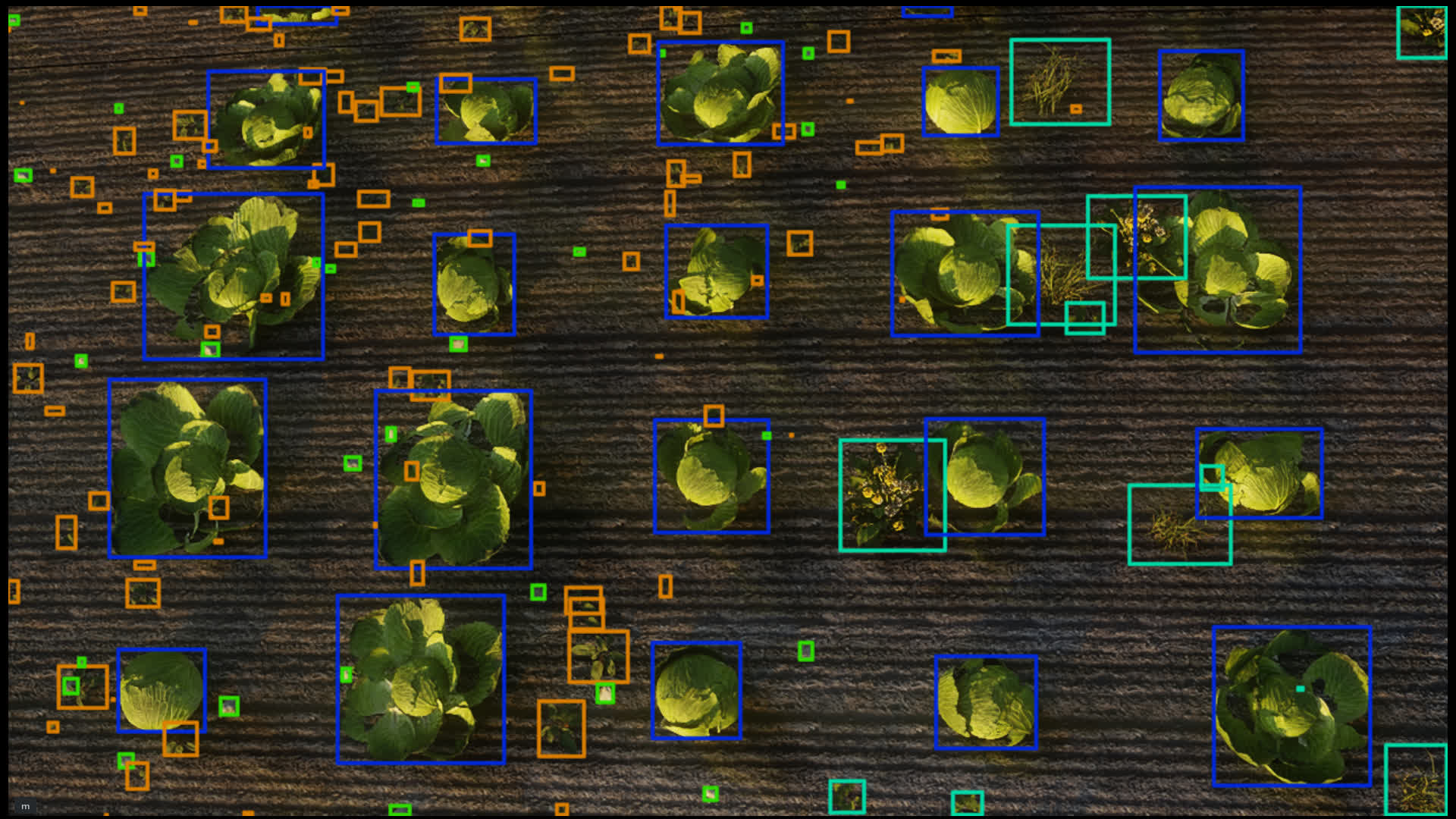}
    \caption{Bounding Box}
  \end{subfigure}
  \vspace{1em}
  \begin{subfigure}[t]{0.45\linewidth}
    \centering
    \includegraphics[width=\linewidth]{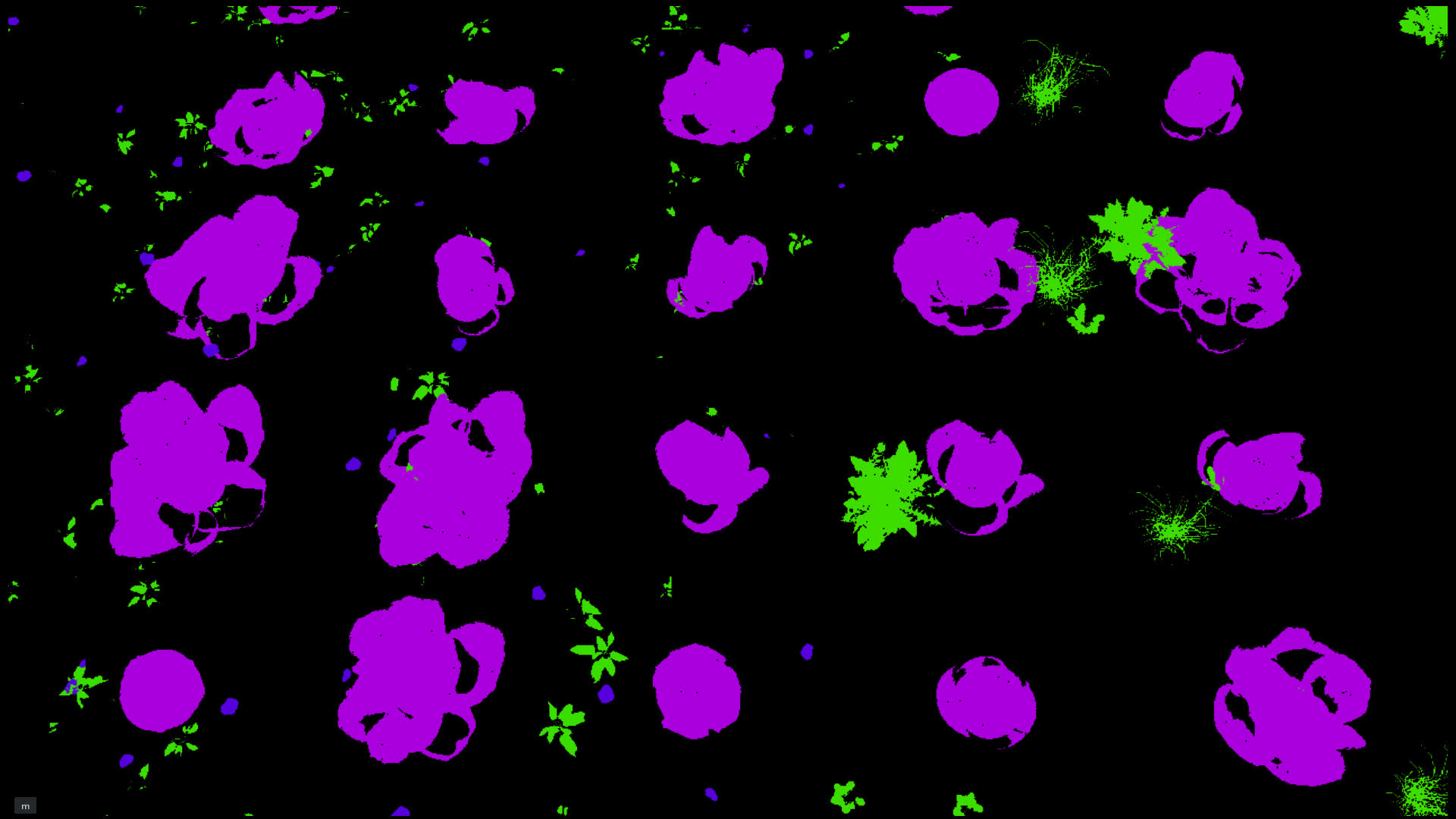}
    \caption{Semantics}
  \end{subfigure}
  \hfill
  \begin{subfigure}[t]{0.45\linewidth}
    \centering
    \includegraphics[width=\linewidth]{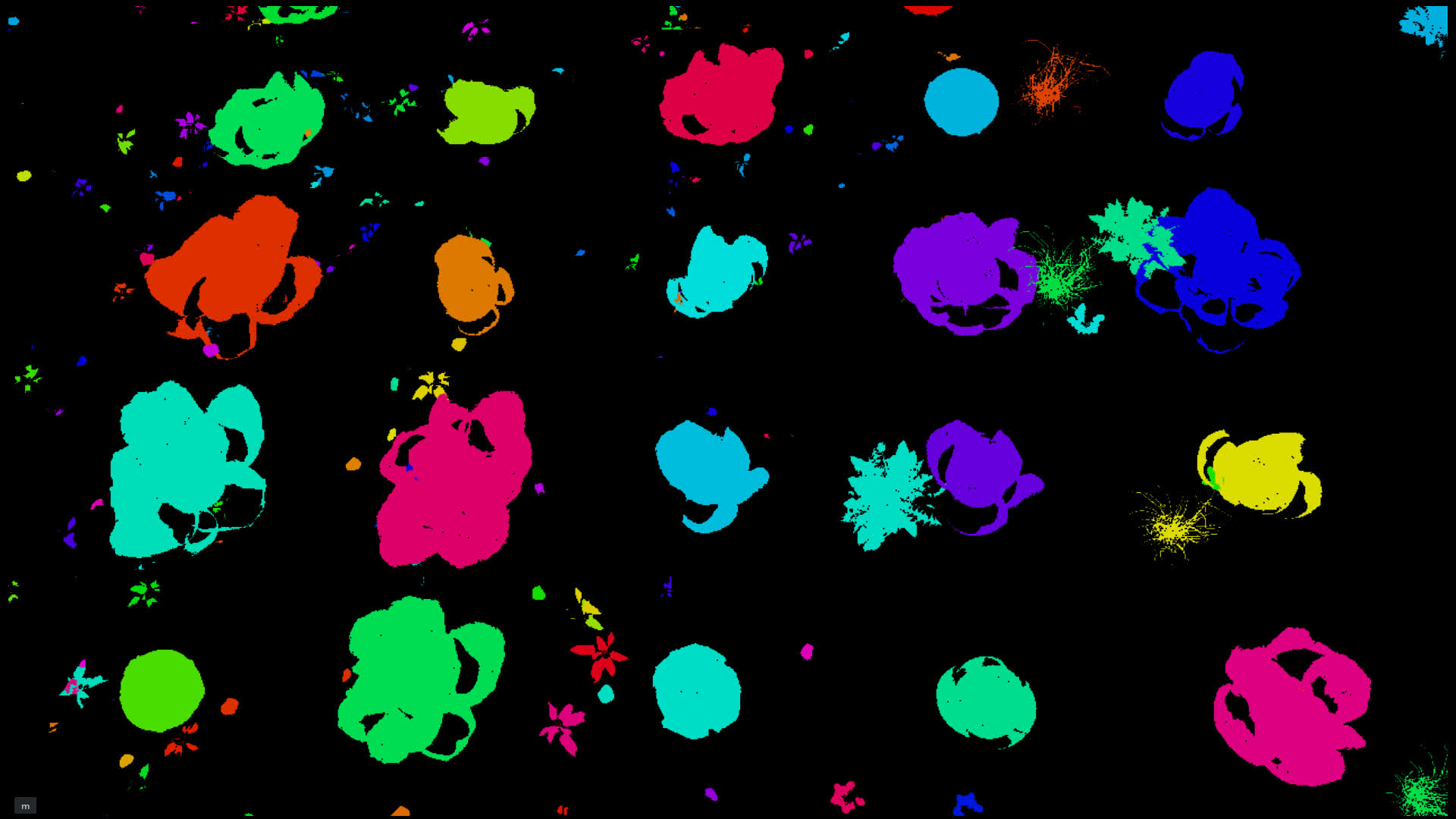}
    \caption{Instance}
  \end{subfigure}
  \caption{Example annotations generated by our framework on a cabbage field image, including bounding boxes, semantic labels, and instance labels for crops and weeds.}
  \label{fig:sim_annotations}
    \vspace{-0.4cm}
\end{figure}

\subsection{Simulated Sensors and ROS Integration}
\label{sec:exp1} 
Our framework procedurally generates photorealistic agricultural environments and publishes sensor data from the simulation as ROS topics. As illustrated in ~\figref{fig:rviz_data}, the simulator provides synchronized multi-modal sensor outputs from the robot’s egocentric view. The example shows the rendered RGB camera frame, its corresponding semantic annotation and LIDAR pointcloud. We also support publishing other data annotations such as bounding boxes or instance labels. \figref{fig:sim_annotations} shows an example of an RGB image along with three generated annotations. In addition, AgriGen is capable of simulating other modalities such as depth, IMU and GPS.

\subsection{Performance Evaluation}
\label{sec:exp2}

\begin{table}[htb] \small
\centering
\begin{tabular}{lccc}
\toprule
Terrain Size
& \shortstack{Workstation\\ Xeon \\RTX 2080 Ti}
& \shortstack{Desktop \\ Intel i9 \\RTX 3080}
& \shortstack{Workstation\\ Ryzen 9 \\2x RTX 3090}\\
\midrule
$50^2 \,\mathrm{m}^2$    & $\approx$~22.14 $\pm$ 4 & $\approx$~41.32 $\pm$ 3 & $\approx$~59.21 $\pm$ 1 \\
$500^2 \,\mathrm{m}^2$   & $\approx$~22.70 $\pm$ 3 & $\approx$~40.00 $\pm$ 2 & $\approx$~58.74 $\pm$ 1 \\ %8793 active objects
$5000^2 \,\mathrm{m}^2$  & $\approx$~22.41 $\pm$ 3 & $\approx$~39.61 $\pm$ 2 & $\approx$~60.78 $\pm$ 2 \\
$50000^2 \,\mathrm{m}^2$ & $\approx$~22.07 $\pm$ 4 & $\approx$~40.93 $\pm$ 2 & $\approx$~60.12 $\pm$ 1 \\
\bottomrule
\end{tabular}
\caption{Simulation performance in frames per second (FPS) of our framework across varied terrain sizes on different machines.}
\label{tab:performance}
\vspace{-0.2cm}
\end{table}

We evaluate the scalability and efficiency of our framework for large-scale environment simulation. As shown in Table~\ref{tab:performance}, all three tested hardware platforms ran the simulation successfully, with FPS scaling with GPU capability. Our streaming architecture loads a fixed number of tiles as the robot moves, so performance depends on local scene complexity rather than total environment size, with only minor FPS fluctuations as the robot's viewpoint changes.

\begin{figure}[h!]
  %\begin{minipage}[b]{.6\linewidth}
    \centering
    \includegraphics[width=0.85\linewidth]{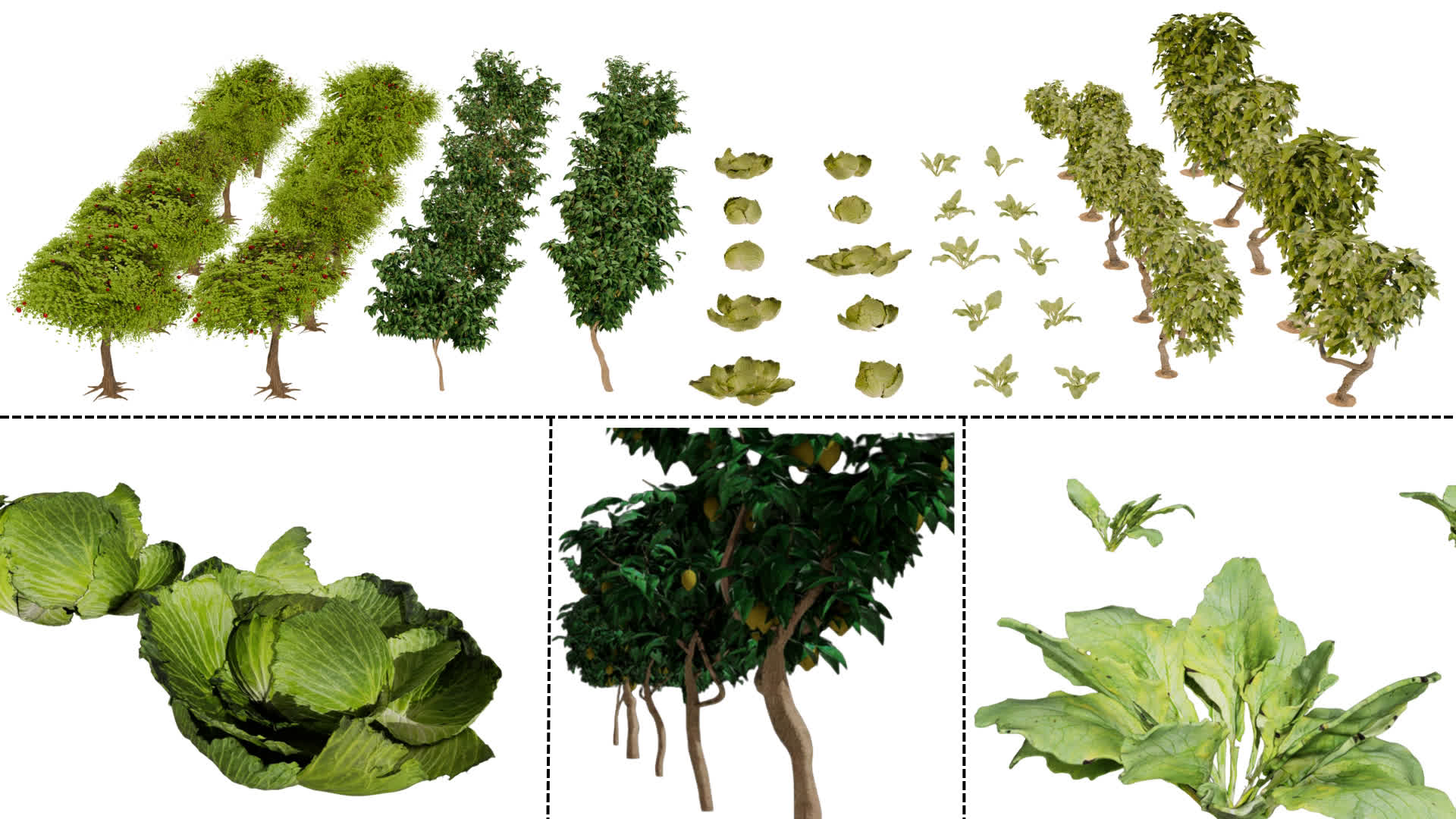}
    \captionof{figure}{Crop assets from the Agrigen dataset. (top) Ten variations across five crop types: apple trees, lemon trees, cabbages, sugar beets, and grapevines. (bottom) Close-up views of cabbages, lemon trees, and sugar beets.}% \caption{Figure caption}
      \label{fig:crops}
  %\end{minipage}\hfill
  %\begin{minipage}[b]{.4\linewidth}
  %  \centering
  \vspace{0.5em}
  \small
  \begin{tabular}{lc}
    \toprule
    Vegetation Type & Variations \\
    \midrule
    \multicolumn{2}{l}{\textbf{Crops}} \\
    \makecell[l]{\textit{Citrus limon} (Lemon Tree)}     & 10  \\
    \makecell[l]{\textit{Malus domestica} (Apple Tree)}  & 10  \\
    \makecell[l]{\textit{Brassica oleracea} (Cabbage)}   & 10   \\
    \makecell[l]{\textit{Beta vulgaris} (Sugar Beet)}    & 10  \\
    \makecell[l]{\textit{Vitis} (GrapeVine)}   & 10 \\
    \midrule
    \multicolumn{2}{l}{\textbf{Weeds}} \\
    \textit{Rumex}               & 5 \\
    \makecell[l]{\textit{Chelidonium majus} \\ } & 5 \\
    \midrule
    \multicolumn{2}{l}{\textbf{Grass and Shrubs}} \\
    \textit{Urtica dioica} & 1 \\
    \textit{Taraxacum}     & 1 \\
    Meadow                 & 1 \\
    Orchidaceae            & 1 \\
    \textit{Pinus}         & 1 \\
    \bottomrule
  \end{tabular}
    \captionof{table}{AgriGen Asset Statistics}
  \label{tab:veg-stats}
    \vspace{-0.4cm}
  %\end{minipage}
\end{figure}

\subsection{Assets}
Next, we demonstrate the visual and geometric fidelity of our plant assets and show that the framework can natively generate row crops, orchards, and vineyards with high realism. 
~\figref{fig:crops} illustrates representative examples of the ten model variations implemented for each crop type, ranging from apple and lemon trees to cabbages and sugar beets, and shows close-up views of their fine geometric detail. This dataset, detailed in ~\tabref{tab:veg-stats}, is released open-source with AgriGen. However, the framework is not limited to these models, and can easily be extended.

\subsection{Domain Randomization}
\label{sec:randomization}

\begin{figure*}[t]
  \centering
  \includegraphics[width=0.98\linewidth]{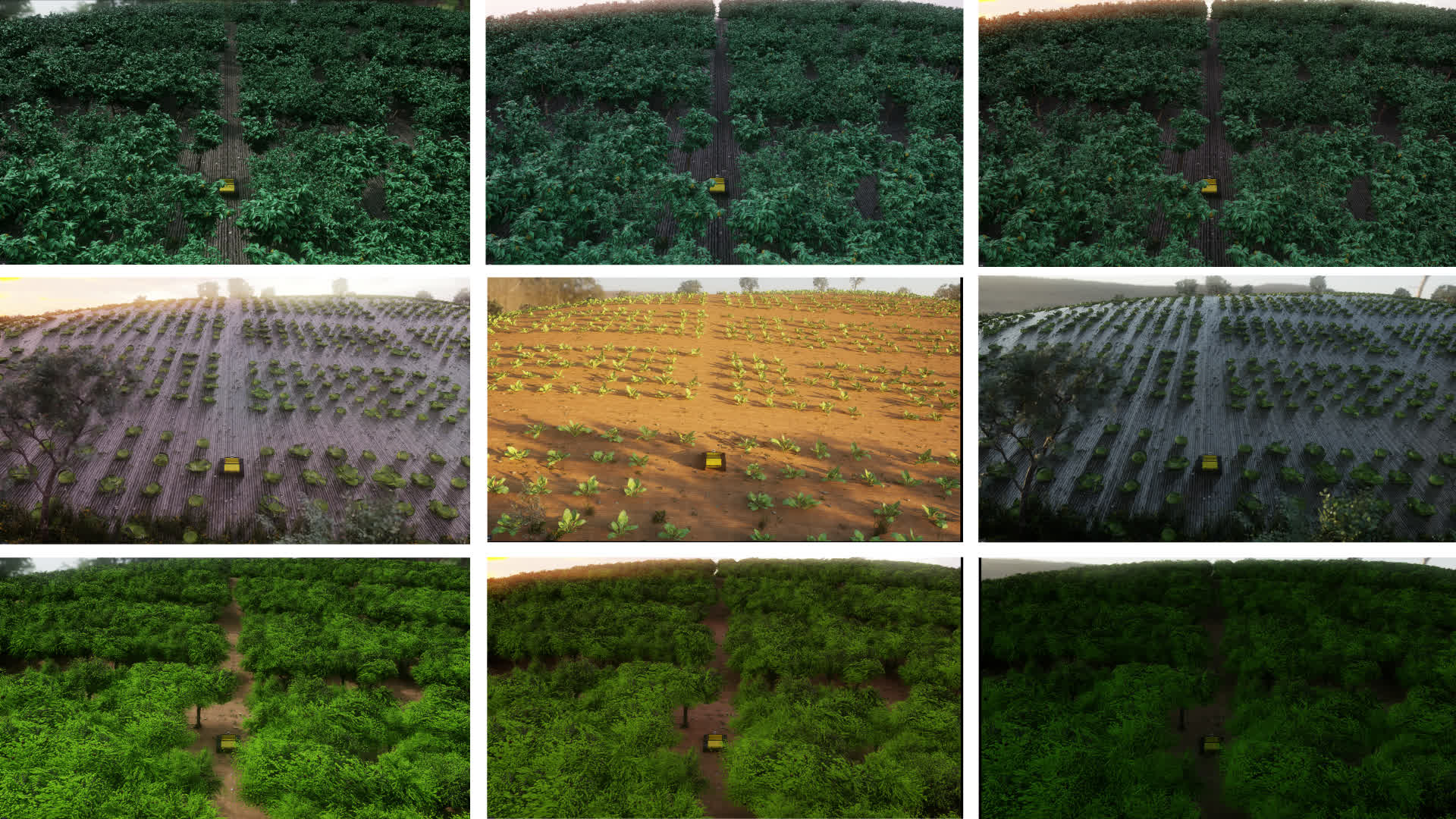}
  \caption{Our framework supports domain randomization across terrain geometry and type, asset geometry, asset type, asset configuration, lighting, and physics parameters such as friction. The examples show: (top row) lemon orchards with randomized terrain geometry, ground textures, and lighting; (middle row) row crops such as cabbages and sugar beets with varied ground textures, vegetation density, and lighting; and (bottom row) apple orchards with different canopy coverage under varied lighting.}
  \label{fig:randomization}
  \vspace{-0.4cm}
\end{figure*}
Finally, we demonstrate the ability of the framework to produce diverse in-field conditions commonly encountered by agricultural robots through domain randomization.
Our framework supports randomization across terrain geometry and type, asset geometry, asset type, asset configuration, environmental illumination, and physics parameters such as friction. As illutrated in~\figref{fig:randomization}, we systematically vary terrain geometry and type, crop and tree geometry, planting patterns, and vegetation density. We also randomize surface friction, ground textures, and illumination to emulate a broad range of field conditions. The figure shows, for example, lemon orchards rendered with distinct terrain morphologies and lighting, crop rows of cabbages and sugar beets under varied soil textures and plant densities, and apple orchards with different canopy coverage and time-of-day illumination.

\section{Conclusion}
\label{sec:conclusion}
In this paper, we presented AgriGen, a novel framework for large-scale agricultural robotics simulation. 
Our framework combines procedural terrain instantiation, vegetation distribution, and streaming-based asset management to generate photorealistic environments at scale. Additionally, AgriGen provides synchronized multi-modal sensor data, including RGB and depth images, image annotations, point clouds, and IMU and GPS measurements, as ROS~2 topics. We show that AgriGen achieves competitive frame rates on consumer-grade hardware while preserving photorealism in large-scale environments. We further showed that the framework delivers high asset realism through our detailed crop models. Finally, we qualitatively demonstrated the domain randomization capabilities of AgriGen over terrain, vegetation, textures and lighting; enabling simulation of diverse in-field conditions for downstream robotics tasks. AgriGen enables researchers in agricultural robotics to systematically and reproducibly evaluate their systems across diverse, large-scale agricultural environments through controlled variation of field conditions and ready access to ground-truth annotations.

\bibliographystyle{plain_abbrv}
\bibliography{2026_Bajpai_ICRA_AgWS-AgriGen}
\end{document}